\documentclass{article}
\usepackage{spconf,amsmath,graphicx, amsfonts, url}
\usepackage{ctable}
\usepackage{tabularx,threeparttable}

\newcolumntype{Y}{>{\centering\arraybackslash}X}

\title{MPE4G : Multimodal Pretrained Encoder for Co-Speech Gesture Generation}
\name{Gwantae Kim$^{1}$, Seonghyeok Noh$^{1}$, Insung Ham$^{1}$ and Hanseok Ko$^{1}$}
\address{$^{1}$School of Electrical Engineering, Korea University, Seoul, South Korea}

\begin{document}
\ninept
\maketitle
\begin{abstract}
When virtual agents interact with humans, gestures are crucial to delivering their intentions with speech. Previous multimodal co-speech gesture generation models required encoded features of all modalities to generate gestures. If some input modalities are removed or contain noise, the model may not generate the gestures properly. To acquire robust and generalized encodings, we propose a novel framework with a multimodal pre-trained encoder for co-speech gesture generation. In the proposed method, the multi-head-attention-based encoder is trained with self-supervised learning to contain the information on each modality. Moreover, we collect full-body gestures that consist of 3D joint rotations to improve visualization and apply gestures to the extensible body model. Through the series of experiments and human evaluation, the proposed method renders realistic co-speech gestures not only when all input modalities are given but also when the input modalities are missing or noisy. The project page is available here\footnote{\url{https://github.com/GT-KIM/Co-speech_gesture_generation}}
\end{abstract}
\begin{keywords}
co-speech gesture generation, pre-trained encoder, multi-modal, neural networks
\end{keywords}
\section{Introduction}
\label{sec:intro}
{\let\thefootnote\relax\footnotetext{The authors of Korea University were supported by DMLab (Q2128981). Corresponding Author:Hanseok Ko.}}
Virtual agents are being deployed to various fields such as the video industry, service industry, news, and social network services\cite{conti2022virtual}. In order to create a human-like behavior, the current virtual agent is played by a real person or uses pre-generated speech and gestures. Instead, to create a more realistic, automatic virtual human, research in various fields such as character design, speech understanding/synthesis\cite{park2017subspace, park2017acoustic, kim2021specmix}, natural language understanding/generation, and gesture\cite{shon2011motion, yang2011modeling} is required. In the present paper, we focus on full-body gestures that occur with text and speech, known as Co-speech gesture generation. Such Co-speech gestures are a representative example of nonverbal communication between people\cite{yoon2020speech}. When gestures are mixed, a much more natural conversation becomes possible in the human-human interaction process compared to standing still during speaking.

In previous works, the rule-based methods\cite{cassell1994animated, cassell2001beat, marsella2013virtual} are proposed to solve the task with a one-to-one mapping between speech and unit gesture pairs. These approaches required a huge amount of data to generate gestures in general scenarios. Recently, the co-speech gesture generation algorithms are progressively improved by the proposals of neural networks-based approaches.\cite{yoon2019robots, yoon2020speech, liu2022learning, qian2021speech}. However, such methods fail to generate the gestures when some parts of the input modalities are corrupted. Therefore, the problem of generalizability remains.

One of the main challenges of co-speech gesture generation studies is determining how to select and encode speech and other modalities. Previous deep learning models designed encoders for every modality and merge their information with concatenation or Recurrent Neural Networks(RNN) manner. However, since the weights of the input modalities are all the same if using these methods, the generator could refer to unusable information when some input modalities are missing or noisy. To solve the problem, we propose the multi-head self-attention\cite{vaswani2017attention} encoder. The proposed encoder module can attend only useful information with attention weights.

Another main obstacle of co-speech gesture generation studies is body model and visualization. Previous Co-speech gesture datasets\cite{cao2019openpose} and methods\cite{yoon2019robots, yoon2020speech, yang2012gesture, yang2017continuous} often use the upper-body or 3D joint position. However, full-body representation with 3D joint rotation is necessary for real applications, such as virtual agents and social robots. Moreover, lower body movements, such as the movement of the center of gravity, increase naturalness. Therefore, we use a 3D joint rotation-based full-body model to represent gestures.

In this paper, we propose a co-speech gesture generation model that uses the text and speech modality. We first propose the embedding and generating model for training joint embedding space between pose, text, and speech. Next, we design a \textbf{Multimodal Pre-trained Encoder for Gesture generation(MPE4G)} that trains with the BERT-style self-supervised learning\cite{devlin2018bert}. Finally, we propose an end-to-end fine-tuning method with the Seq2Seq model. Our contributions can be summarized as follows:
\begin{itemize}
    \item We propose the neural networks model with the multi-head self-attention-based encoder for robust co-speech gesture generation. Our model outperforms baselines in quantitative and qualitative aspects.
    \item We propose a self-supervised pre-training method to train the proposed encoder model.
    \item We present the 3D joint rotation-based full-body representation to represent gestures. With the representation, It can provide rich visualization and can be applied to real applications.
\end{itemize}

\section{Proposed method}
In this section, we present the body model and neural networks model for 3D Co-speech gesture generation. In summary, we used the SMPL-X\cite{SMPL-X:2019} body model to capture more detailed motion and provide realistic visualization. We propose a neural networks model and its training strategy to generate co-speech gestures using text or speech. The proposed model has the advantage of robust generation when some modality is not provided.

\begin{figure}[t] 
\begin{center}
\includegraphics[width=0.7\linewidth]{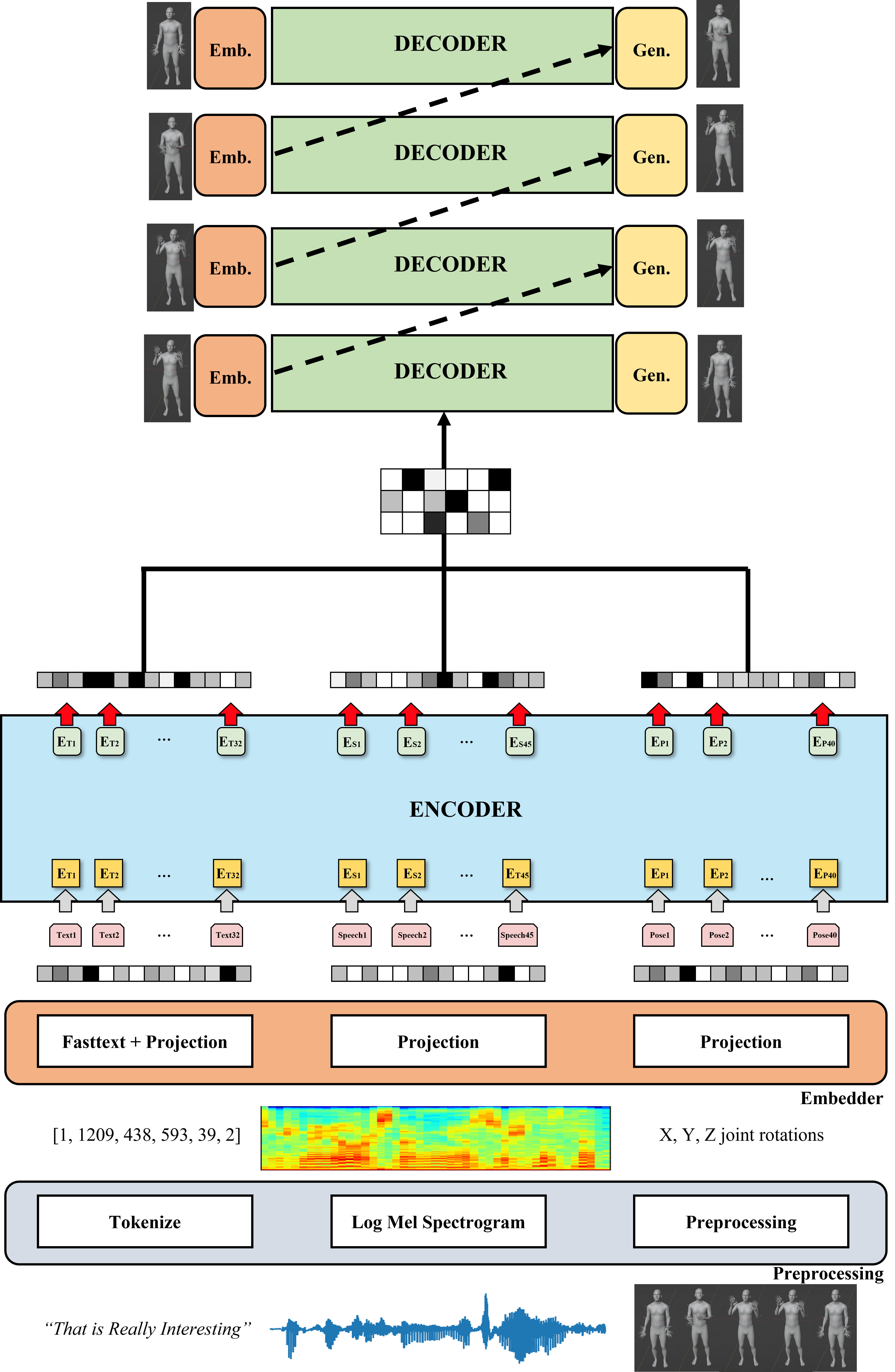}
\end{center}
\caption{Illustration of the model pipeline.}
\label{fig : neural networks}
\end{figure}

\begin{figure}[t] 
\begin{center}
\includegraphics[width=0.7\linewidth]{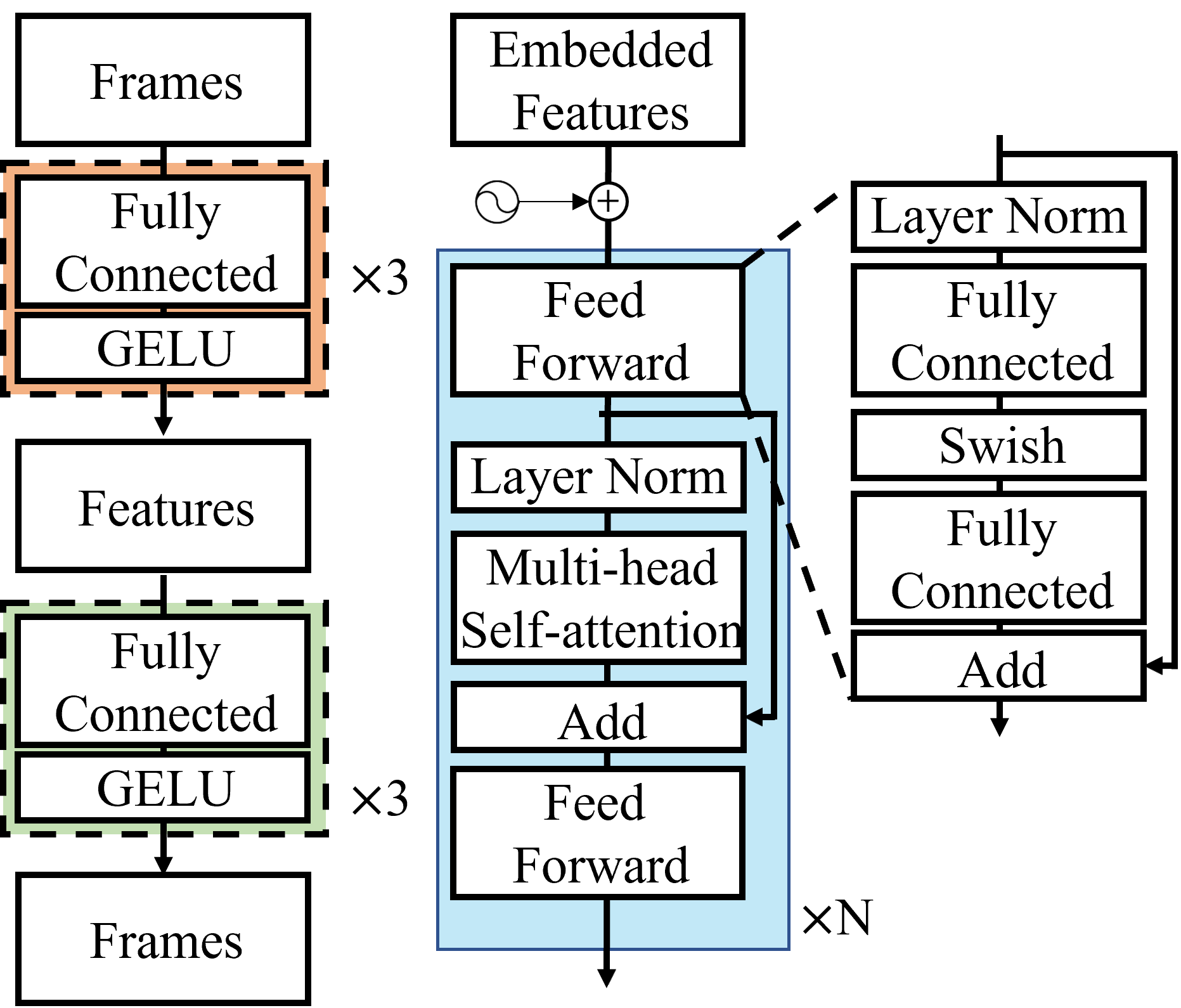}
\end{center}
\caption{Detailed Model structure. Left: embedder and generator. Right: encoder block.}
\label{fig : model structure}
\end{figure}

\subsection{Body Model}
The previous works about co-speech gesture generation\cite{yoon2019robots, yoon2020speech, liu2022learning} mainly used the TED gesture dataset, which applied Openpose\cite{cao2019openpose} and 3D pose estimator\cite{pavllo20193d} to TED videos from Youtube. The body is represented by 2D or 3D joint positions in the dataset. Despite their successful co-speech gesture generation results, the joint position-based representation is not suitable to control the 3D body because of bone length and rigging problems. In the recent work\cite{liu2022learning}, they tried to apply Expose\cite{choutas2020expose} to TED videos and represent the body using unit directional vectors with mean bone lengths. It can solve the problems of bone length and rigging. However, they used upper-body joints only and did not visualize their results with a 3D human model. We used the SMPL-X\cite{SMPL-X:2019} body model and our own human body model to solve the problems of the joint position-based representation and provide more rich full-body visualization.


\subsection{Neural Network Model}

\noindent\textbf{Motivations} The challenges for designing an encoder are "How to integrate multi-modal inputs as hidden representations", and "How to transform text/speech domain to pose domain by the semantic relationship". The fully-connected embedder initially reduces the domain gap between different modalities with joint embedding loss(stage 1). Next, the pre-trained multi-head self-attention-based encoder generates integrated hidden representations with self-supervised learning(stage 2). Thanks to the properties of the self-attention mechanism that focuses on important features, more rich hidden representations can be acquired. We also designed a decoder with multi-head self-attention layers, instead of RNN layers. Since frames of the RNN layers are too dependent on the just previous frame. In contrast, frames of the Transformer layers can refer to all other frames, as well as focus on important frames. Therefore, we found that the transformer can generate motions robustly.

\noindent\textbf{Preprocessing} The time-aligned data is cropped to 1.33 seconds long. We build a word-level dictionary using the dataset to tokenize the text and the text is zero-padded. The speech is converted into log-Mel spectrogram with Fourier transform parameters [nfft, win\_length, hop\_length, n\_mels] = [2048, 60ms, 30ms, 128]. The pose vector consists of normalized 3D joint rotation angles in radians. The shapes of the preprocessed features are text $t$=[b, $l_T$], speech $s$=[b, $l_S$, 128], and pose $p$=[b, $l_P$, 165], where b denotes batch size and $l$ denotes feature length. In this paper, $l_T$, $l_S$, $l_P$ set to 32, 45, 40.

\noindent\textbf{Stage 1} In the first stage, the frame-wise embedder and generator are trained with joint embedding loss and reconstruction loss. The preprocessed text features are projected using Fasttext\cite{bojanowski2016enriching} word embedding and additional 3 fully-connected layers as shown in the left of Fig.2. The preprocessed speech and pose features are projected using 3 fully-connected layers. The generator consists of 3 fully-connected layers for each modality, and its role is to reconstruct outputs from embedded features.

Let embedded features of text, speech, and pose are $E_{T}$, $E_{S}$, and $E_{P}$, respectively. The joint embedding loss is calculated by text-speech and text-pose alignment without zero-padded text, which is calculated by (1) and (2). The text-speech joint embedding loss is 
\begin{equation}
    \mathcal{L}_{J_{S}} = \sum_{N_{t}=1}^{l_{T}}|E_{T_{N_{t}}} - \frac{1}{l_{ST}}\sum_{N_{s}=l_{ST}N_{t}}^{l_{ST}(N_{t}+1)}E_{S_{N_{s}}}|
\end{equation}
where $l_ST$ = $l_S / l_T$. Remark that $l_T$ is non-zero-padded text length, not 32 in stage 1. Similar, the text-pose joint embedding loss is
\begin{equation}
    \mathcal{L}_{J_{P}} = \sum_{N_{t}=1}^{l_{T}}|E_{T_{N_{t}}} - \frac{1}{l_{PT}}\sum_{N_{p}=l_{PT}N_{t}}^{l_{PT}(N_{t}+1)}E_{P_{N_{p}}}|
\end{equation}
where $l_PT$ = $l_P / l_T$. The L1 distance between text embedding, speech embedding, and pose embedding at the same timestamps is decreased with (1) and (2). The reconstruction loss is calculated with cross-entropy loss and L1 distance.
\begin{equation}
    \mathcal{L}_{recon} = \mathcal{L}_{CE}(t, \hat{t}) + |s - \hat{s}| + |p - \hat{p}|.
\end{equation}
Finally, the embedder and generator are optimized with loss $\mathcal{L} = 0.01\cdot(\mathcal{L}_{J_{S}} + \mathcal{L}_{J_{P}}) + \mathcal{L}_{recon}$, Adam optimizer, learning rate 0.005, batch size 64, and 50 epochs.

\noindent\textbf{Stage 2} In the second stage, the multimodal encoder is trained with self-supervised learning. The input feature of the multimodal encoder is the concatenation of the text, speech, and pose features. The multimodal encoder has N=4 attention layers that consist of two feed-forward layers and one multi-head self-attention layer with residual connections. Moreover, the sinusoidal positional encoding gets added at the beginning of the encoder. Motivated by masked token prediction and next-word prediction from BERT\cite{devlin2018bert}, we propose self-supervised learning methods. The input text is fully ignored with 10\% prob., one word is masked with 72\% prob., one word is changed to a random word with 9\% prob., or no masking with 9\% probability. Similarly, the input speech is fully ignored with 10\% prob., 5 continuous frames are masked with 72\% prob., 5 continuous frames are changed to random values with 9\% prob., or no masking with 9\% probability. The input pose is fully ignored with 10\% prob., 5 continuous frames are masked with 9\% prob., 5 continuous frames are changed to random values with 9\% prob., and the last 30 frames are masked for the next frame prediction with 63\% or no masking with 9\% probability.
The multimodal encoder reconstructs texts, speeches, and poses using masked inputs with reconstruction loss (3). The masked inputs pass through the embedder, encoder, and generator in that order to estimate the reconstructed outputs. The encoder model can learn relations and joint-embedded space for texts, speeches, and poses. Finally, the encoder is optimized with reconstruction loss $\mathcal{L} = \mathcal{L}_{recon}$, Adam optimizer, learning rate 0.005, batch size 64, and 100 epochs. Note that the weights of the embedder and generator are frozen in this stage.

\noindent\textbf{Stage 3} In the third stage, the embedder, encoder, decoder, and generator are jointly trained with supervised learning. The decoder model consists of an N=1 multi-head attention block, which is shown on the right of Fig. 2, but changed multi-head self-attention into multi-head attention.
The decoder generates each frame with the auto-regressive method in training and testing, instead of the teacher-forcing method. When the decoder model predicts the n-th gesture frame, the query of the multi-head attention is previous frames, and the key and value of the multi-head attention are all output features of the encoder. To preserve the consistency of the frames, we used 10 pre-poses, which are outputs of previous timestamps. When $n < 10$, the input of the model is picked from pre-poses, instead of generated frames.
The reconstruction losses of the text and speech in stage 2 are also used in stage 3 because we want to preserve information on each modality in the encoder outputs. Moreover, we used pose loss\cite{yoon2019robots}, which contains L1 reconstruction loss, motion velocity loss, and motion variance loss. We used similar settings from \cite{yoon2019robots}, but we maximize motion velocity loss, instead of minimizing, it to generate more active movements. Finally, we trained the embedder, encoder, decoder, and generator with reconstruction loss, pose loss, Adam optimizer, learning rate 0.005, batch size 32, and 360 epochs.

\begin{figure}[t] 
\begin{center}
\includegraphics[width=0.7\linewidth]{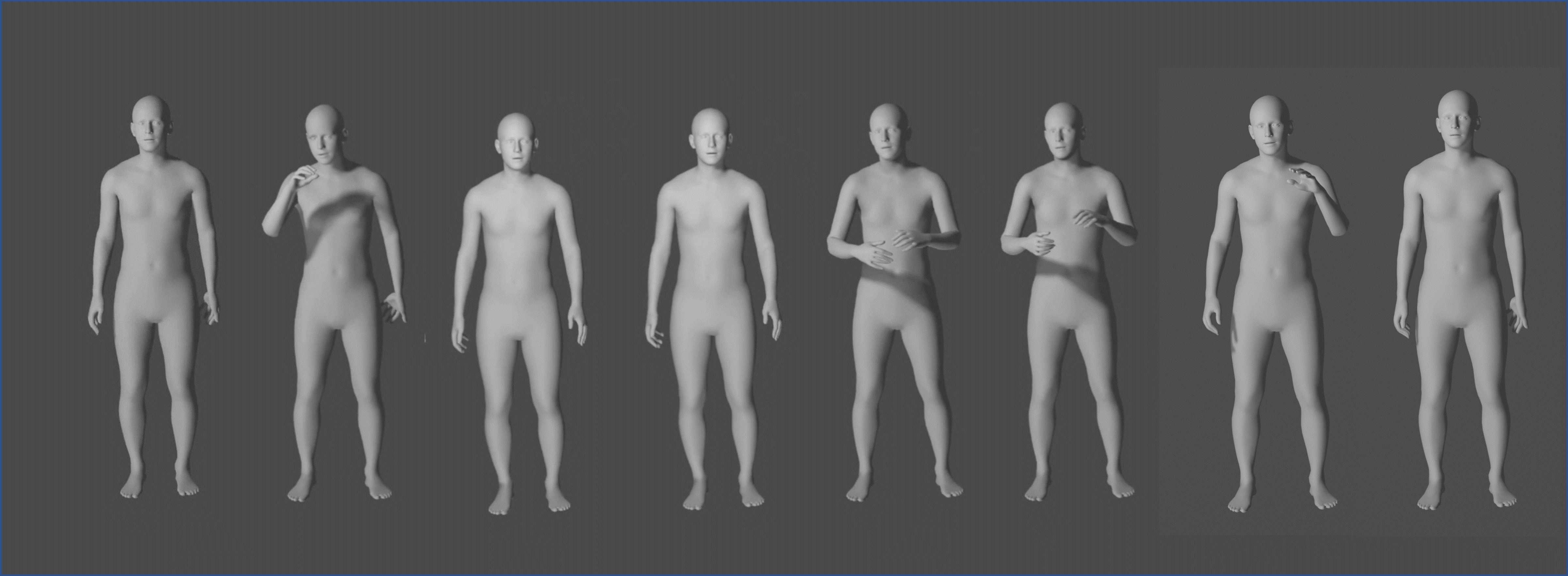}
\end{center}
\caption{Visualization results of the generated gestures.}
\label{fig : output samples}
\end{figure}

\section{Experiments}

\begin{table*}[t]
\caption{The quantitative results on GENEA-EXPOSE dataset. Input T is text and input S is speech. The uparrow denotes that higher score is better and downarrow is opposite.}
\label{table:ex1}
\centering
    \begin{tabular}{c|c|c|c|c|c|c|c|c}
    \specialrule{.2em}{.1em}{.1em}
    Models & Input & $MPJAE\downarrow$ & $MMD\downarrow$ & $FGD\downarrow$ & Diversity$\uparrow$ & BC$\uparrow$ & FLOPs & infer. time \\ \hline
    Noisy GT & - & 0 & 0 & 2.63 & 157.47 & 0.930 & - & -  \\
    Attention Seq2Seq\cite{yoon2019robots} & T & \textbf{0.073} & 0.395 & 7.46 & 82.47 & 0.268 & 1.99M & 0.021s\\
    Trimodal\cite{yoon2020speech} & T & 0.089 & 0.865 & 18.06 & 27.33 & 0.862 & 8.87M & 0.036s \\
    HA2G\cite{liu2022learning} & T & 0.127 & 0.263 & 12.31 & 186.75 & 0.828 & 57.7M & 0.083s \\
    \textbf{MPE4G(Ours)} & T & 0.106 & 0.327 & 9.67 & 133.81 & 0.593 & 91.21M & 0.040s \\ \hline
    Speech2Gesture\cite{ginosar2019gestures} & S & 0.075 & 0.841 & 16.303 & 21.894 & 0.818 & 0.11M & 0.053s \\
    Trimodal\cite{yoon2020speech} & S & 0.106 & 0.327 & 9.67 & 134.17 & 0.593 & 8.87M & 0.036s \\
    HA2G\cite{liu2022learning} & S & 0.129 & 0.270 & 11.077 & \textbf{206.55} & 0.864 & 57.7M & 0.083s \\
    \textbf{MPE4G(Ours)} & S & 0.098 & 0.23  & 5.46 & 125.90 & 0.686 & 91.21M & 0.040s \\ \hline
    Trimodal\cite{yoon2020speech} & T+S & 0.102 & 0.297 & 12.40 & 120.76 & 0.682 & 8.87M & 0.036s \\
    HA2G\cite{liu2022learning} & T+S & 0.130 & 0.271 & 10.90 & 201.55 & \textbf{0.879} & 57.7M & 0.083s  \\
    SEEG\cite{liang2022seeg} & T+S & 0.091 & 0.623 & 16.336 & 52.51 & 0.233 & 1.17M & 0.1s \\
    \textbf{MPE4G(Ours)} & T+S & 0.095 & \textbf{0.188} & \textbf{4.96} & 127.24 & 0.682 & 91.21M & 0.040s \\
    \specialrule{.2em}{.1em}{.1em}
    \end{tabular}
\end{table*}

\subsection{Dataset}
\noindent\textbf{GENEA challenge 2022 Dataset}
We mainly used the GENEA challenge 2022\cite{yoon2022genea} dataset, which is based on the Talking With Hands 16.2M gesture dataset\cite{lee2019talking}. The GENEA challenge dataset was split into a training set (18 h), a validation set (40 min), and a test set (40 min). The training set and the validation set contain motion data(.bvh format), aligned text data, and aligned speech data. The test set only contains text data and speech data. We trained the model with a training set and tested the model with a validation set. We did not use the test set because the ground truth gestures do not exist.

Since we use the SMPL-X body model unlike the GENEA challenge, we first converted the motion data format from bvh to the SMPL-X body model. We generated videos by applying the challenge visualization toolkit to motion data and used Expose\cite{choutas2020expose} to each video frame-by-frame. The frame-wise Expose results are merged and used as Co-speech gesture data. Since the timestamps of each frame are the same as the video, we can use aligned text and speech files easily. Finally, we successfully collect aligned full-body gestures, which contain not only upper body joints but also hand and lower body joints, audio, and text samples. The new data is named as \textbf{GENEA-EXPOSE}. Since output poses extracted by Expose contain noise, especially around hand poses, the ground truth gestures are not clean. So we set the ground truth name as "Noisy GT". We will reduce the noise using better 3D pose estimation in future works.

\subsection{Evaluation metrics}

\noindent\textbf{Mean Per Joint Angle Error (MPJAE)}
Mean Per Joint Angle Error(MPJAE)\cite{ionescu2013human3} is widely used to measure the difference between ground truth pose and generated pose. Since our data structure contains 3D joint angles, the MPJAE can be simply implemented with mean absolute error.

\noindent\textbf{Maximum Mean Discrepancy}
The Maximum Mean Discrepancy(MMD) metric measures the similarity between two distributions. It is used to measure the quality of generated samples compared with ground truth in deep generative model\cite{zhao2019self} and human action generation\cite{kim20223d, yu2020structure}. The metric has also been applied to evaluate the similarity between generated actions and the ground truth in \cite{wang2020learning}, which has been proven consistent with human evaluation. We used MMDavg settings from \cite{kim20223d} for evaluation.

\noindent\textbf{Fr\'{e}chet Gesture Distance (FGD)}
Fr\'{e}chet Gesture Distance (FGD) is a kind of inception-score measurement. It measures how close the distribution of generated gestures is to the ground truth gestures. FGD is
calculated as the Fr\'{e}chet distance between the latent representations of real gestures and generated gestures. We train an auto-encoder on the GENEA-EXPOSE dataset, similar to \cite{liu2022learning, yoon2020speech}, to calculate latent representations.

\noindent\textbf{Beat Consistency Score}
The Beat Consistency(BC) score is a metric for motion-audio beat correlation, which is frequently used in the dance generation. We used the BC to observe consistency between audio and generated pose. We used the implementation written by \cite{liu2022learning}, which is optimized for the co-speech gesture generation task.

\noindent\textbf{Diversity}
Diversity measures how well the model can generate variate motions. We used similar settings to \cite{liu2022learning}. We used the FGD auto-encoder to get latent feature vectors of the generated gestures and calculate the average feature distance. The number of synthesized gesture pairs is 500, which is randomly selected from the test set.

\subsection{Quantitative Evaluation}
The quantitative results are shown in Table 1. The attention Seq2Seq\cite{yoon2019robots} model with text inputs has good MPJAE and FGD scores, but other scores are bad. These results mean the model sometimes generates ground-truth-like gestures, but it cannot generate natural gestures. Trimodal\cite{yoon2020speech} and HA2G\cite{liu2022learning}, and MPE4G(ours) used both speech and text modality. The proposed MPE4G framework outperforms baselines and state-of-the-art models in MMD and FGD, which measure distribution similarity. In detail, MMD and FGD determine the overall generation performance of the generative models because they compare overall distributions between GT and generated samples, not sample by sample. MPE4G did not have the best score on MPJAE, Diversity, and BC scores. However, the balance between these scores is more important than each individual score. MPJAE compares joint angles of ground truth and generated samples for every test sample. Therefore, low MPJAE means the generated sample is exactly the same as the ground truth. However, not only the exactness but also the diversity of the generative models are important. Since MPJAE cannot measure the diversity of the generated samples, we measured Diversity and BC. We expect high Diversity and low BC. However, if the generated motions are meaningless but have high variation, the Diversity goes high(high Diversity but unwanted results). Moreover, BC focuses on beat consistency with the speech signal. If there is motion in the part where there is an audio signal, the BC comes out high. If the meaningless motions are repeated whether speech exists or not, BC goes high(low BC but unwanted results). Therefore, low MPJAE, high Diversity, and low BC must happen at once. We think MPE4G has the best-balanced scores on MPJAE, Diversity, and BC. Moreover, the state-of-the-art models produce worse results when some input modalities are missing, but the proposed models can preserve generation performance in the same case.

\subsection{Qualitative Evaluation}
The qualitative results of the GENEA-EXPOSE dataset are summarized in Table 2. We used the Mean Opinion Score(MOS) about naturalness, smoothness, and synchrony. A higher score denotes better results. The proposed method MPE4G outperforms state-of-the-art methods.
Especially, the generated gestures using MPE4G achieve better results compared to noisy ground truth although the models trained by the noisy ground truth. Therefore, the proposed method promises better results when training data is corrupted.

The visualization results of the generated samples are shown in Fig. 3. We can provide full-body visualization with a 3D body model. More samples can be found on the project page.

\subsection{Limitations and Future works}
The generated samples are too dependent on the co-speech dataset since the model is trained by L1-like loss, instead of adversarial loss. To solve the problem and generate more abundant poses we will use the prior-pose dataset in the self-supervised stage in future works. Moreover, the dataset can be used to train discriminators for adversarial training. We also plan to reduce noise in the GENEA-EXPOSE data using other 3D pose estimation methods and improve methods to generate various motions.

\begin{table}[t]
\caption{The qualitative results on GENEA-EXPOSE dataset}
\label{table:ex2}
\centering
    \begin{tabular}{c|c|c|c}
    \specialrule{.2em}{.1em}{.1em}
    Models & Naturalness & Smoothness & Synchrony \\ \hline
    Noisy GT & 1.958 & 1.955 & 2.719 \\
    Trimodal & 3.326 & 3.587 & 2.154 \\
    HA2G & 1.675 & 1.585 & 1.915 \\
    \textbf{MPE4G(ours)} & \textbf{3.457} & \textbf{3.629} & \textbf{3.241} \\
    \specialrule{.2em}{.1em}{.1em}
    \end{tabular}
\end{table}

\section{Conclusion}
We presented a multi-head self-attention-based encoder and decoder for 3D co-speech gesture generation from text and speech. Our model not only outperformed state-of-the-art methods but could generate gestures when some input modalities are missing or corrupted. We collected a new dataset, named GENEA-EXPOSE, for full-body representation, better visualization, and usage in real applications. Through a series of experiments, the proposed method successfully proved its contributions. We also provided rich visualization tools using the SMPL-X body model. In our future work, we plan to reduce noise in the GENEA-EXPOSE data using other 3D pose estimation methods and improve methods to generate various motions. Moreover, we will collect large and clean motion datasets to improve generation performance.

\bibliographystyle{IEEEbib}
{\footnotesize\bibliography{ref}}

\end{document}